\documentclass[runningheads]{llncs}

\usepackage[mobile]{eccv}

\usepackage{eccvabbrv}
\usepackage{stfloats}
\usepackage{float}
\usepackage{multirow}
\usepackage{mathtools}
\usepackage{bbding}
\usepackage{bm}
\usepackage{amsfonts,amssymb}
\usepackage{algorithm,algorithmicx,algpseudocode}
\usepackage{colortbl}
\definecolor{mygray}{gray}{.9}
\definecolor{mypink}{rgb}{.99,.91,.95}
\definecolor{mycyan}{cmyk}{.3,0,0,0}
\usepackage{graphicx}
\usepackage{booktabs}

\usepackage[accsupp]{axessibility}  % Improves PDF readability for those with disabilities.

\usepackage[pagebackref,breaklinks,colorlinks,citecolor=eccvblue]{hyperref}
\usepackage{orcidlink}

\begin{document}

\title{StoryImager: A Unified and Efficient Framework for Coherent Story Visualization and Completion}

\titlerunning{StoryImager}

\author{Ming Tao\inst{1,2} \and
Bing-Kun Bao\inst{1,2} \and
Hao Tang\inst{3} \and \\
Yaowei Wang\inst{2} \and
Changsheng Xu\inst{2,4,5}
}

\authorrunning{M.~Tao, BK.~Bao et al.}

\institute{Nanjing University of Posts and Telecommunications \and
Peng Cheng Laboratory \and  Robotics Institute, Carnegie Mellon University  \and
MAIS, Institute of Automation, Chinese Academy of Sciences  \and
School of Artificial Intelligence, University of the Chinese Academy of Sciences 
}

\maketitle

\begin{abstract}
Story visualization aims to generate a series of realistic and coherent images based on a storyline. 
Current models adopt a frame-by-frame architecture by transforming the pre-trained text-to-image model into an auto-regressive manner.
Although these models have shown notable progress, there are still three flaws.
1) The unidirectional generation of auto-regressive manner restricts the usability in many scenarios.
2) The additional introduced story history encoders bring an extremely high computational cost. 
3) The story visualization and continuation models are trained and inferred independently, which is not user-friendly.
To these ends, we propose a bidirectional, unified, and efficient framework, namely StoryImager. 
The StoryImager enhances the storyboard generative ability inherited from the pre-trained text-to-image model for a bidirectional generation. 
Specifically, we introduce a Target Frame Masking Strategy to extend and unify different story image generation tasks. 
Furthermore, we propose a Frame-Story Cross Attention Module that decomposes the cross attention for local fidelity and global coherence. 
Moreover, we design a Contextual Feature Extractor to extract contextual information from the whole storyline. 
The extensive experimental results demonstrate the excellent performance of our StoryImager.
Code is available at \url{https://github.com/tobran/StoryImager}.

\keywords{Generative Model \and Story Visualization \and Story 
  Completion}
\end{abstract}

\section{Introduction}
\label{sec:intro}

The last few years have witnessed the great success of large pre-trained generative models for a variety of applications.
Among them, text-to-image synthesis is one of the important tasks of generative models. 
Due to its practical value, text-to-image synthesis has become an active research area, leading to the development of large pre-trained text-to-image autoregressive and diffusion models, \eg, DALL-E~\cite{ramesh2021zero} and LDM~\cite{rombach2022high}.
Based on the powerful image-generative ability, some recent work \cite{storydalle, rahman2023make, pan2022synthesizing} tried to extend pre-trained text-to-image models beyond generating a single image to generating a sequence of story images, a task referred to as story visualization.

Unlike single text-to-image generation, story visualization aims to generate coherent and visually appealing story images based on a sequence of story descriptions. 
The task requires the model to capture the essence of the story and the relationships between different story descriptions and then translate them into realistic images that tell a story in a coherent manner.
Some work expanded this task by introducing story continuation to generate subsequent story images based on both given story descriptions and historical frames.
The importance of story visualization and continuation lies in its potential applications in entertainment, education, and multimedia storytelling.

\begin{figure}[t] \small
    \centering
    \includegraphics[width=\linewidth]{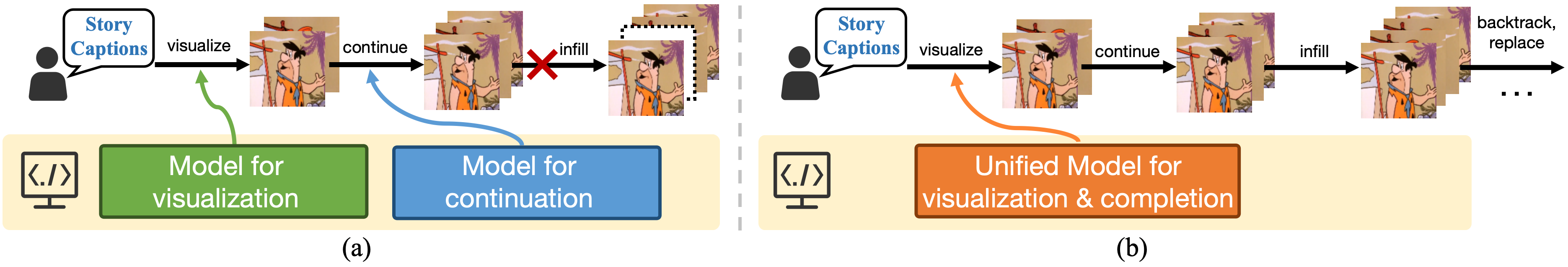}
    \caption{(a) Existing models adopt the auto-regressive generative approach, which restricts usability in many scenarios. And the users need to switch between models to meet their current requirements 
    (b) Our proposed StoryImager unifies different tasks into one model, which is more comprehensive to tackle various generative requirements.}
    \label{fig1}
    \vspace{-0.6cm}
  \end{figure}

\begin{figure*}[t] \small
  \centering
  \includegraphics[width=\linewidth]{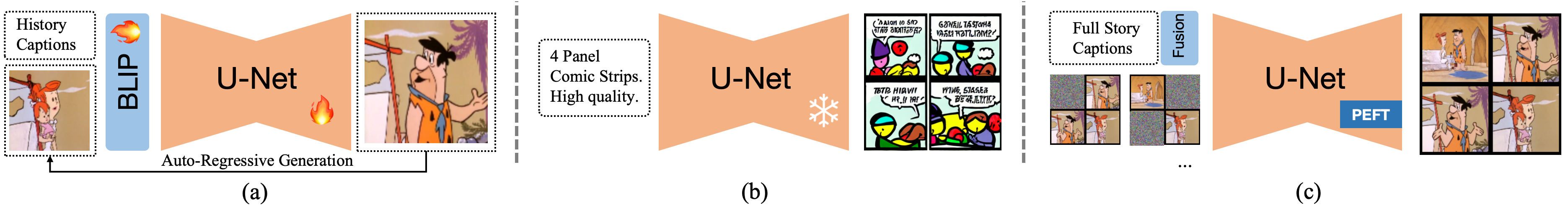}
  \caption{(a) Existing models introduce large models to encode the history information for auto-regressive generation.
           (b) The storyboard generative ability of Stable Diffusion \cite{rombach2022high} learned from pretraining process.
           (c) Our proposed StoryImager inherits the storyboard generative ability and unifies different tasks through a masking strategy. }
  \label{fig2}
  \vspace{-0.6cm}
\end{figure*}

Although impressive results have been presented in previous work \cite{storydalle, rahman2023make, pan2022synthesizing}, they still suffer from three flaws.
First, the capabilities of current models are limited. 
The autoregressive generative architecture they adopted \cite{storydalle,pan2022synthesizing,rahman2023make} only supports frame-by-frame generation due to unidirectional attention. 
It restricts the applicability of models in scenarios that require referencing future or bidirectional frames, such as story backtracking, insertion, and replacement. 
This limitation prevents users from making arbitrary modifications and extensions based on existing story frames.
Second, existing models are trained and inferred independently for different tasks. 
The current story image generation task includes Story Visualization and Story Continuation. 
However, existing methods tackle them as two independent tasks. They train two independent models and infer separately.
Figure \ref{fig1}(a) shows that independent models require users to store, load, and specify different large models for each requirement.
Thus, it increases the complexity of usage and increases hardware requirements.
Third, current models introduce large multimodal models such as BLIP \cite{li2022blip} to encode history story captions and frames, then stack extracted hidden states directly, as shown in Figure \ref{fig2}(a). 
The additional large models and stacked long hidden states enlarge the story model and bring an extremely high computing budget.

To address the above issues, we propose a novel unified story visualization and completion framework named StoryImager. 
For the first issue, we further extend the scope of the tasks and categorize them into story visualization and completion based on the different given conditions.
Story visualization is only conditioned on story captions, whereas story completion is conditioned on both story captions or images provided at any arbitrary frame.
The new scope fulfills the synthesis requirements and makes the task more comprehensive.
Under the new scope, we propose a Storyboard-based Generation (Storyboard-Gen).
As shown in Figure \ref{fig2}(b), the large pretraining enables the stable diffusion to generate plausible storyboards. 
The Storyboard-Gen inherits the pre-trained ability and enhances it for story visualization and completion through Parameter-Efficient Fine-Tuning (PEFT) \cite{peft}. 
The Storyboard-Gen enables the bidirectional synthesis of story images in the storyboard.
As shown in Figure \ref{fig1}(b), it expands our model in the story image generation task and allows it to be applied in a broader range of scenarios.

For the second issue, we introduce a Target Frame Masking Strategy (Target-Mask) for training and inference that addresses both story visualization and completion tasks.
As shown in Figure \ref{fig2}(c), during training, the frames are completely masked for story visualization and partially masked randomly for story completion.
During inference, users simply need to load a single model and mask the frames they want to synthesize, enabling them to handle various requirements seamlessly without reloading other models.
The Target-Mask provides a more flexible generation approach to meet the specific generation goals of users.

For the third issue, we design a decomposed Frame-Story Cross Attention Module (Frame-Story CAM).
It leverages the inherent cross-attention module of the pre-trained diffusion model and decomposes it into local frame-level and global story-level cross-attention. 
This decomposed design ensures both the image quality of each frame and the overall consistency of the story images.
Furthermore, the self-attention in the storyboard enables visual feature interactions between frames. 
Thus, our model can effectively fuse story caption and visual features without additional large multimodal encoders.

Overall, our contributions can be summarized as follows:
\begin{itemize}
    \item We extend the task of story image synthesis and propose a unified story visualization and completion framework to synthesize high-fidelity and coherent story images.
    \item We propose a Storyboard-based Generation, which enables bidirectional synthesis of story images in the storyboard.
    \item We introduce a Mask-based training and inference strategy that consolidates various synthesis requirements into masked story image prediction.
    \item We design a Frame-Story Cross Attention Module, which ensures the image fidelity of individual frames and the overall coherence of the entire story.
    \item Extensive qualitative and quantitative experiments on two challenging datasets demonstrate that the proposed StoryImager outperforms existing models.
\end{itemize}

\section{Related Work}
\subsection{Text-to-Image Synthesis}
Text-to-image generative models greatly impact story synthesis models and are always adopted as the generative backbone.
Contemporary advances in text-to-image synthesis have primarily centered around three primary frameworks: Generative Adversarial Networks, Auto-regressive models, and Diffusion models. 
% GANs, encompassing AttnGAN \cite{xu2018attngan}, DM-GAN \cite{zhu2019dm}, DF-GAN \cite{tao2020df}, and GALIP \cite{tao2023galip}, 
Text-to-Image GANs \cite{zhang2017stackgan, xu2018attngan, zhu2019dm, tao2020df, tao2023galip} employ adversarial training strategies between generators and discriminators. 
Large-scale autoregressive models, such as DALL·E \cite{ramesh2021zero}, Make-A-Scene \cite{gafni2022make}, and Parti \cite{yu2022scaling}, 
have exhibited commendable scalability and exceptional proficiency in synthesizing images. 
Diffusion models \cite{sohl2015deep,dhariwal2021diffusion,ho2020denoising,ho2022cascaded,nichol2021improved}, 
including VQ-Diffusion \cite{gu2022vector}, GLIDE \cite{nichol2021glide}, DALL-E2 \cite{ramesh2022hierarchical}, Latent Diffusion Models (LDM) \cite{rombach2022high}, and Imagen \cite{saharia2022photorealistic}, 
have garnered significant interest within the research community. 
As likelihood-based models, they effectively avoid the common pitfalls of mode-collapse and potential instability during training that are often associated with GANs, 
facilitating the generation of a more diversified assortment of images. 

\subsection{Story Visualization and Continuation}

StoryGAN \cite{storygan} first proposes the story visualization task and provides a GAN-based sequence generation model consisting of a deep RNN-based context encoder and two discriminators for images and stories. 
Subsequently, some works follow and refine this network. 
For example, CP-CSV \cite{song2020character} introduces a foreground segmentation generation module to optimize the consistency of characters and backgrounds in the story. 
Both DUCO \cite{ducostorygan} and VLC \cite{vlcstorygan} enhance semantic consistency by dual learning. Still, the former also considers inter-image sequence consistency via copy-transform, while the latter focuses more on textual information via introducing external common-sense information to complement textual details. 
Word-Level SV \cite{wordlevelsv} and Clustering GAN \cite{li2022clustering} simplify the two-level GAN network of StoryGAN \cite{storygan} and further optimize story quality through word-level fine-grained features and clustering learning, respectively. 
In addition to work based on GAN models, 
VQ-VAE-based VP-CSV \cite{vpcsv}, pre-trained model DALL-E-based StoryDALL-E \cite{storydalle}, and diffusion model-based AR-LDM \cite{pan2022synthesizing} can also tackle story visualization. 
Notably, to improve task generalization, StoryDALL-E \cite{storydalle} defines a new task story continuation, which introduces the source frame to generate unseen plots and characters. 
AR-LDM \cite{pan2022synthesizing} and Make-A-Story \cite{rahman2023make} transform the diffusion model into an autoregressive story image generative manner to synthesize image sequence.
The AR-LDM introduces the large pre-trained BLIP to encode history stories and frames, and the Make-A-Story proposes a visual memory module that captures the context information.
There are some works \cite{gong2023talecrafter,zhu2023cogcartoon,liu2023intelligent} that focus on open-ended story visualization.
TaleCrafter \cite{gong2023talecrafter} proposes an interactive story visualization by incorporating sketch and layout controls.
The CogCartoon \cite{zhu2023cogcartoon} proposes a character-plugin generation to alleviate dependence on data and storage.
The Intelligent Grimm \cite{liu2023intelligent} collects a diverse open-ended story dataset from YouTube and E-books.

Our proposed StoryImager differs greatly from the previous story visualization and continuation models.
It extends existing tasks that can tackle various generative tasks, such as story visualization, story continuation, story infilling, and story backtracking. 
Our model adopts a Storyboard-based Generation approach to enable bidirectional story image generation. 
Compared to previous models, our StoryImager is more effective and convenient in synthesizing high-quality and coherent story images.

\section{The Proposed Method}

In this work, we expand the scope of story image generation and propose a unified and contextually coherent framework. 
This framework effectively addresses both the story visualization and completion tasks simultaneously, providing comprehensive coverage for the generation process.
To synthesize high-fidelity and coherent story images under different tasks, we propose:
(\romannumeral1) a Storyboard-Gen approach that enables bidirectional synthesis of story images in a storyboard.
(\romannumeral2) a Target-Mask strategy for training and inference that consolidates diverse synthesis requirements into masked story image prediction.
(\romannumeral3) a Frame-Story CAM with a Context Feature Extractor (Context-FE) ensures both the visual fidelity of individual frames and the overall coherence of the story.
In the following of this section, we first present a comprehensive overview of our StoryImager. 
Following that, we provide detailed explanations of the proposed Storyboard-Gen, Target-Mask, FrameStory-CAM, and Context-FE in detail.

\subsection{Model Overview}
\begin{figure*}[t] \small
    \centering
    \includegraphics[width=\linewidth]{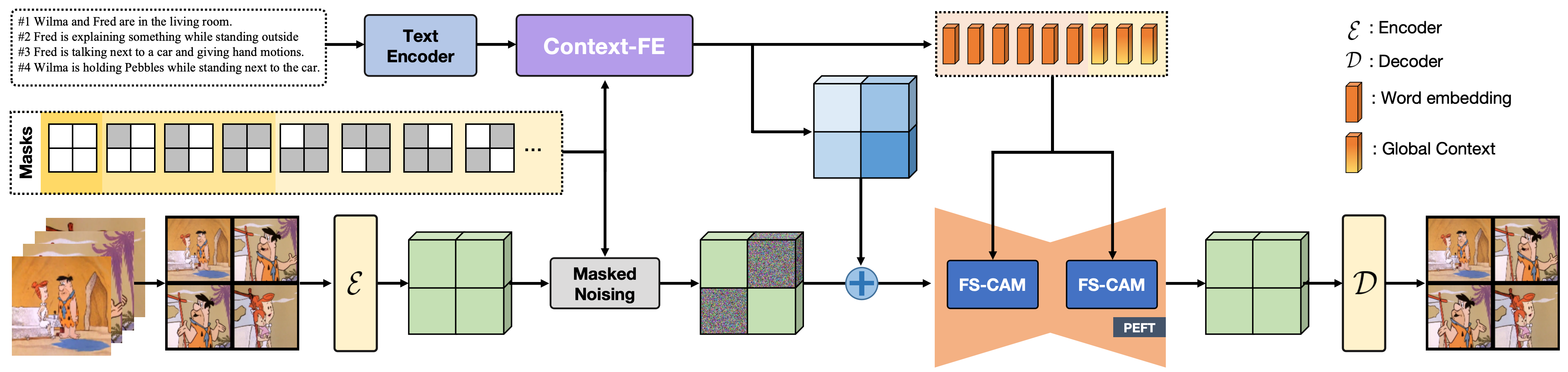}
    \caption{The architecture of StoryImager for story visualization and completion. Our StoryImager adopts a Storyboard-based Generation approach to enable bidirectional story image generation. It unifies different tasks through the Target Frame Masking Strategy.}
    \label{fig3}
    \vspace{-0.4cm}
  \end{figure*}

As illustrated in Figure \ref{fig3}, our proposed StoryImager consists of a pretrained Text Encoder, a pair of Image Encoder $\mathcal{E}$ and Decoder $\mathcal{D}$ of pretrained autoencoder, 
a Contextual Feature Extractor (Context-FE), a pretrained diffusion Model \cite{rombach2022high} with Parameter-Efficient Fine-Tuning (PEFT) \cite{peft}, 
a Target Frame Masking Strategy, and Frame-Story Cross Attention Modules (Frame-Story-CAM).
The story images are first encoded into latent space, and the text encoder encodes the text descriptions into word embeddings.
The Context-FE extracts global contextual information and predicts a frame-aware latent prior.
The Target Frame Masking Strategy masks target story images for training and inference.
Then the U-Net of the pretrained diffusion model takes the latent features, word embeddings, and contextual information as input and
then fuses them through the Frame-Story Cross Attention Module at each layer. 
The whole model is trained by predicting the noise of the masked parts.
Finally, after reversing the diffusion process multisteps, the latent features are decoded to the targe images.

\subsection{Storyboard-based Generation}

Existing story image generation models adopt an auto-regressive architecture for synthesizing multiple images from story captions. 
They transform the pre-trained text-to-image model into a frame-by-frame generative model, enabling the generation of story image sequences. 
However, the limitation of the auto-regressive architecture lies in its unidirectional generation, which hampers its applicability in various scenarios. 
During the process of creating story images, users may need to modify a specific frame within the story or incorporate new story elements into an existing narrative. 
These requirements necessitate a model that can refer to the content of existing frames bidirectionally, allowing for effective interaction with the existing story.
Nevertheless, unidirectional generation disregards requirements such as story insertion, replacement, and backtracking, significantly constraining the usability of these models in such contexts.

In recent years, large-scale pretrained models have demonstrated remarkable zero-shot capabilities in various natural language processing and computer vision tasks. 
They keep the structure of pre-trained models while transforming the downstream tasks to closely resemble the approaches used in pertaining.
For example, Prompt learning \cite{radford2019language,raffel2020exploring,brown2020language} leverages the construction of suitable prompts or contexts to enable GPT \cite{brown2020language} to perform text classification tasks and specific text generation \cite{brown2020language,gao2020making,liu2022p}. 
By setting up downstream tasks to closely resemble the pertaining process, we can better utilize the knowledge acquired during pretraining.
Inspired by this, we explore the potential of large pretrained diffusion models in generating storyboards. 
As shown in Figure \ref{fig2}(b), we find that the pretrained diffusion model is able to synthesize plausible storyboards. 
Thus, a question arises: Can we transform the task of story visualization and completion into the generation of storyboard images? 
Motivated by it, we introduce Storyboard-based Generation (Storyboard-Gen) as our story image synthesis approach.
The Storyboard-Gen arranges the story images in sequential order and places them into panels, forming a storyboard image.
In contrast to the auto-regressive generative approach, where diffusion models would need to learn story generation from scratch.
Our storyboard-based Generation leverages the inherent storyboard image generative ability of pretrained models and trains them in a manner that aligns closely with their original pretraining objectives. 
Furthermore, compared with the auto-regressive generative approach, our Storyboard-Gen enables the bidirectional synthesis of story images in a storyboard.
It forms the foundation for constructing a unified framework for this task.

\subsection{Target Frame Masking Strategy}

Previous story image generative models focus on story visualization and continuation. 
Story continuation is a variant of story visualization that shares the same goal but incorporates a source frame (i.e., the first frame) to guide the generation of subsequent frames.
However, the information in a single source frame is limited, and such a setup restricts the application scenarios of the model.
Moreover, existing methods tackle story visualization and continuation as two independent tasks. 
They train two independent models separately.
This not only increases the training burden, but also introduces complexity for users, 
who must constantly switch between models to meet their specific requirements when generating long story images.
Therefore, we extend the story image generative task and introduce the Target Frame Masking Strategy (Target-Mask) for training and inference to unify story visualization and completion.

In the forward process, the Target-Mask randomly samples a binary mask $m$ to indicate the desired frames for generation. 
In the case of story visualization, the mask consists entirely of True, indicating that all frames are target generation frames. 
For story completion, only the frames requiring completion are marked True in $m$, while the rest of the frames remain unmasked. 
This strategy enables us to generate specific frames based on current requirements selectively.
Then the Target-Mask applies a masked noising process on the latent of storyboard $x$.
We define $x_0=x$ and only add noise on the latent of masked frames instead of the whole latent:
\begin{align}
    &\tilde{x}_t = \sqrt{\bar{\alpha}_t} x_0 + \sqrt{1-\bar{\alpha}_t}\epsilon\\ \nonumber
    &x_t = \tilde{x}_t \odot m + x_0 \odot (1-m),
\end{align}
where $\epsilon\sim \mathcal{N}(\textbf{0}, \textbf{I})$ and $t$ is the timestep in the forward process.
% We use $x_t$, $m$, and $c$ as input to the model 
The Target-Mask enables the model to utilize visual information in given story frames and learn to recover masked frames $x_0\odot m$.
This ensures that generated frames in the mask $m$ are consistent with the given frames. 
Following \cite{ho2020denoising} we train a network $\epsilon_{\theta}$ to predict the noise $\epsilon$ from the noisy $x_t$:
\begin{align}
    \mathcal{L}_{\text{DM}} = \mathbb{E}_{ \epsilon \sim \mathcal{N}(0, I)}\left[\|m \odot\epsilon - m \odot \epsilon_\theta(x_t, t, s)\|_2^2\right].
\end{align}
where $s$ is the story caption.
The typical training scheme used in stable diffusion is predicting the loss on the whole latent. 
Since our target is to predict masked story frames according to given frames and story captions, we only calculate the loss on masked frames.

In the inference phase, we apply a masked noising process on the target frames in storyboard $x_T=\epsilon \odot m + x_0 \odot (1-m)$, where $T$ is the number of sampling steps. 
The unmasked frames are kept from the denoising process at each step.
Then we reverse the diffusion process and obtain the completed storyboard $x_0$.

By incorporating Target-Mask, our approach enhances the flexibility and adaptability of the model for both story visualization and completion tasks. 
It allows us to generate complete storyboards or fill in missing frames seamlessly, based on the provided context and requirements.
Compared with previous methods, our masking strategy unifies different tasks in one framework and extends the ability of the story image generative model.
It simplifies the training complexity and user interaction process, ensuring a streamlined and user-friendly experience in creating coherent story images.

\subsection{Frame-Story Cross Attention Module}

\begin{figure*}[t] \small
  \centering
  \includegraphics[width=\linewidth]{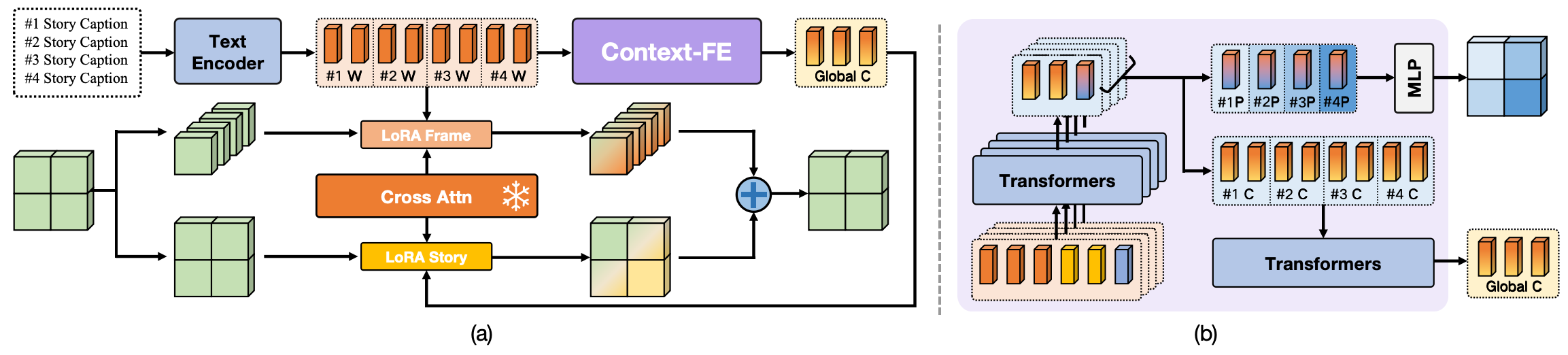}
  \caption{(a) The architecture of Frame-Story Cross Attention Module. It decomposes the cross-attention module into story-level and frame-level cross-attention to enable local image fidelity and global story coherence. 
           (b) The proposed Contextual Feature Extractor summarizes the whole text information, extracts global contextual information, and predicts the frame-aware latent prior for the U-Net. }
  \label{fig4}
  \vspace{-0.4cm}
\end{figure*}

To ensure that the generated story images in the storyboard possess both realistic visual content and overall narrative coherence, 
we introduce a novel module called Frame-Story Cross Attention Module (Frame-Story CAM). 
As illustrated in Figure \ref{fig4}(a), the Frame-Story CAM decomposes the pretrained cross-attention modules of the diffusion model in each layer into two components: 
the cross-attention mechanism between local frames and the corresponding word embeddings, 
and the cross-attention mechanism between the entire storyboard and the extracted contextual story features.

At the frame level, the cross-attention mechanism divides the encoded latent features into four smaller frame-specific latent features at each layer, 
aligning with the sequence of frames in the storyboard.
Then these frame-specific latent features are flattened to the batch dimension.
Simultaneously, the word embeddings extracted from each story caption are also arranged in the order of frames and flattened to the batch dimension.
Finally, we fuse the flattened frame-specific latent features and their corresponding word embeddings through a cross-attention mechanism.
This process incorporates the textual information from each story caption into the visual features of each local story frame, 
thereby facilitating seamless alignment between the visual and textual information for each local story frame.

At the storyboard level, the cross-attention mechanism performs feature fusion between the entire encoded latent features and the global contextual features extracted by the Contextual Feature Extractor (Context-FE). 
This fusion ensures consistency between the overall visual representation of the story images and the global storylines.

In particular, we did not introduce separate cross-attention mechanisms from scratch for the frame and storyboard levels. 
Instead, we adopted the Low-Rank Adaptation (LoRA) \cite{hu2021lora}, 
which freezes the pretrained model weights and introduces trainable rank decomposition matrices into each layer. 
Furthermore, both frame-level and storyboard-level cross-attention at each layer utilizes the pretrained parameters of the current layer's cross-attention.
They learn two sets of trainable low-rank decomposition matrices for local and global cross-modal feature fusion.
This design of two low-rank decompositions from one pretrained layer allows our Frame-Story CAM to inherit the excellent cross-modal fusion capabilities learned from the pretraining for both local and global fusion,
while significantly reducing the number of trainable parameters. 

Armed with the proposed Frame-Story CAM, our model can maintain the visual quality of individual frames and ensure the overall coherence of the story.
The specifically designed LoRA applied on the frame level and the storyboard level cross attention strikes a balance between model adaptability and parameter efficiency, 
resulting in an effective and efficient framework for generating high-fidelity and coherent story images.

\subsection{Contextual Feature Extractor}

Story captions often comprise multiple sentences, and concatenating all of the captions into a single long text can easily exceed the maximum length of the pretrained text encoder. 
Moreover, the pretrained text encoder is not specifically trained for the combination of multiple sequential story captions, 
which may result in the loss of sequential information across frames.
To effectively extract the contextual information of the entire story, we propose the Contextual Feature Extractor (Context-FE).

As shown in Figure \ref{fig4}(b), Context-FE receives word embeddings from each story caption and incorporates additional learnable query embeddings to summarize the information in the word embeddings of each frame. 
We employ two types of learnable query embeddings: context query embeddings and prior query embeddings. 
These query embeddings are stacked along the length dimension and fed into a network consisting of two layers of Transformers.
By utilizing this network, we obtain context embeddings and prior embeddings of each frame's story caption. 
Then, the prior embeddings are fed into an MLP (Multilayer Perceptron) to predict frame-aware latent prior. 
The frame-aware latent prior is reshaped to match the size of the latent representation, which is $4{\times}64{\times}64$.
Afterward, the frame-aware latent prior is added to the input latent representation. 
This incorporation of the prior information provides the Frame-Story CAM with valuable contextual cues, 
enabling it to better differentiate between different frames. 
Simultaneously, the context embeddings of each frame are concatenated and fed into a context summarizer consisting of two layers of Transformers.
The global context feature summarized from the context summarizer contains the contextual information of the entire story caption. 
Incorporating the global context feature allows the model to effectively leverage the comprehensive story context for coherent story image synthesis.

\section{Experiments}
\label{sec:experiments}

\subsection{Datasets}
We evaluate our approach on two challenging datasets: Pororo-SV \cite{pororo} and Flintstones-SV \cite{flintstones}.
Each story in these two datasets contains five consecutive frames.
The Pororo-SV dataset consists of 10191/2334/2208 samples in the training, validation, and testing sets, respectively. 
The Flintstones-SV dataset consists of 20132/2071/2309 samples in the training, validation, and testing sets, respectively.
The image resolution of these two datasets is $128\times128$.
We resize the images to 248 and pad them with 4 pixels, resulting in a final image size of 256$\times$256. 
The storyboard consists of four-story images with a resolution of 512$\times$512. 
Each image of Pororo-SV and Flintstones-SV corresponds to 1 story description.
The partitioning of these two datasets into training, validation, and testing subsets is conducted in line with the established practices in previous studies \cite{storydalle,pan2022synthesizing,rahman2023make}.

\subsection{Training and Evaluation Details}

Our method is based on the stable diffusion model with publicly available checkpoints v1.5. 
We freeze the pretrained autoencoder and text encoder and finetune the U-Net through LoRA with $\alpha=4$, $r=128$. 
% We select the ``conv_in'', ``to_q'', ``to_k'', ``to_v'' modules for LoRA finetuneing.
We use the AdamW optimizer to train our model.
We set the learning rate $0.001$ for the U-Net and $0.0001$ for the Contextual Feature Extractor. 
All models were trained on 8 $\times$ NVIDIA RTX A6000 GPUs.
The network is trained 300 epochs and 150 epochs on Pororo-SV and Flintstones-SV for about 20 hours.
Following the previous story visualization and continuation works \cite{wordlevelsv,storydalle,pan2022synthesizing,rahman2023make}, we adopt the Fr\'echet Inception Distance (FID) \cite{heusel2017gans} to evaluate the image fidelity of synthesized story images.
We also adopt the Fr\'echet Story Distance (FSD) \cite{song2020character,wordlevelsv} to evaluate the overall quality of the story image sequence.
Different from FID, the FSD considers temporal consistency.
The FID and FSD provide a robust measure to evaluate image fidelity and story consistency of synthesized story image sequences.
During the evaluation, we sample images using the DDIM scheduler~\cite{ddim} for 50 inference steps with a guidance scale set to 6.0 and crop the generated images from the storyboard. 
In addition, we also conduct large-scale human evaluations regarding visual quality, visual consistency, and story relevance.

\subsection{Quantitative Evaluation}
To evaluate the performance of our proposed StoryImager, we compare it with several state-of-the-art story visualization and continuation methods \cite{song2020character,vlcstorygan,wordlevelsv,pan2022synthesizing,storydalle}, which have achieved impressive results. 
The comparison results for Pororo-SV and Flintstones-SV are shown in Table \ref{table1}.
From the table, we can observe that our StoryImager achieves better FID and FSD against other models on both story visualization and continuation tasks.
Our model also surpasses the recently proposed Causal-Story \cite{song2023causal}, which even adopts additional causal reasoning to the AR-LDM framework.
The improvements are more obvious in FSD, which evaluates the overall coherence of the synthesized story images.
The results demonstrate that our model excels at synthesizing high-fidelity and contextually coherent story images.
Furthermore, our model has the ability to perform story completion, a capability that was absent in previous models. 
In the story completion testing, we randomly mask parts of the story image and then predict the missing portions. This testing process does not include story visualization.
We list the performance of StoryImager on the story completion task, which is a unique feature of our model.

To demonstrate the efficiency of our framework, we conducted a comparison with the state-of-the-art AR-LDM framework \cite{pan2022synthesizing}. 
AR-LDM utilizes an autoregressive approach for story image generation and has been adopted as a foundational framework \cite{song2023causal}.
Table~\ref{table3} presents the cost which is evaluated on 8$\times$ and 1$\times$ A6000 with batch size 1 for training and inference, respectively.
We assess resource and time requirements of inference on a single GPU, as it better reflects the usage of most users.
As shown in Table~\ref{table3}, our model achieves improved performance and extended tasks while reducing hardware and time requirements.
The results indicate that our model is a more effective and efficient framework for story visualization and completion.

\newcommand{\CC}[1]{\cellcolor{mygray}}
\begin{table}[t] \small
\centering
\caption{The results of FID and FSD compared with the state-of-the-art models on the test set of Pororo-SV and Flintstones-SV. \XSolidBrush indicates that the model does not support this task.}
\vspace{-0.4cm}
\resizebox{0.99\linewidth}{!}{
\begin{tabular}{lccccccc}
\toprule
\multirow{2}*{Dataset}   &\multirow{2}*{Method}                 & \multicolumn{2}{c}{Story Visualization}     & \multicolumn{2}{c}{Story Continuation}       & \multicolumn{2}{c}{Story Completion}       \\ 
                                                                 \cmidrule(lr){3-4}                            \cmidrule(lr){5-6}                             \cmidrule(lr){7-8}                        
                         &                                      & FID $\downarrow$    & FSD $\downarrow$      & FID $\downarrow$     & FSD $\downarrow$      & FID $\downarrow$    & FSD $\downarrow$          \\ \midrule
                         &StoryGAN \cite{storygan}              & 78.64               & 94.53                 &\XSolidBrush          &\XSolidBrush           & \XSolidBrush        & \XSolidBrush               \\ 
                         &CP-CSV \cite{song2020character}       & 67.76               & 71.51                 & \XSolidBrush         & \XSolidBrush          & \XSolidBrush        & \XSolidBrush        \\
                         &DUCO \cite{ducostorygan}              & 95.17               & 171.70                & \XSolidBrush         & \XSolidBrush          & \XSolidBrush        & \XSolidBrush        \\
                         &VLC \cite{vlcstorygan}                & 94.30               & 122.07                & \XSolidBrush         & \XSolidBrush          & \XSolidBrush        & \XSolidBrush        \\
                         &WL-SV \cite{wordlevelsv}              & 56.08               & 52.50                 & \XSolidBrush         & \XSolidBrush          & \XSolidBrush        & \XSolidBrush        \\ 
\multirow{1}*{Pororo-SV\cite{pororo}}  &StoryDALL-E \cite{storydalle}         & \XSolidBrush        & \XSolidBrush          & 25.90                & 45.70                 & \XSolidBrush        & \XSolidBrush       \\  
                         &MEGA-StoryDALL-E \cite{storydalle}    & \XSolidBrush        & \XSolidBrush          & 23.48                & -                     & \XSolidBrush        & \XSolidBrush                \\   
                         &Make-A-Story\cite{rahman2023make}     & 27.33               & 51.20                 & 22.66                & 44.22                 & \XSolidBrush        & \XSolidBrush               \\ 
                         &\CC{}AR-LDM\cite{pan2022synthesizing} &\CC{}16.59           &\CC{}35.33             &\CC{} 17.40           &\CC{} 37.52            &\CC{} \XSolidBrush   &\CC{} \XSolidBrush             \\   
                         &Causal-Story\cite{song2023causal}     & 16.28               & -                     & 16.98                & -                     & \XSolidBrush        & \XSolidBrush               \\
                         &\CC{}StoryImager (Ours)               &\CC{}\textbf{15.63}  &\CC{}\textbf{28.13}    &\CC{}\textbf{15.45}   &\CC{}\textbf{27.10}    &\CC{}\textbf{14.72}  &\CC{}\textbf{26.77}            \\ 
                         & \\[-2.3ex]        
                         \hline         
                         & \\[-2.3ex]        
                         &StoryGAN \cite{storygan}              & 90.55               & 122.71                &\XSolidBrush          &\XSolidBrush           & \XSolidBrush        & \XSolidBrush      \\ 
                         &StoryGANc \cite{storydalle}           & -                   & -                     & 90.29                & -                     & \XSolidBrush        & \XSolidBrush        \\ 
                         &WL-SV \cite{wordlevelsv}              & 72.37               & 91.30                 & \XSolidBrush         & \XSolidBrush          & \XSolidBrush        & \XSolidBrush        \\ 
\multirow{1}*{Flintstones-SV\cite{flintstones}} &StoryDALL-E \cite{storydalle}         & \XSolidBrush         & \XSolidBrush          & 26.49                & 54.30                 & \XSolidBrush        & \XSolidBrush        \\
                         &MEGA-StoryDALL-E\cite{storydalle}   & \XSolidBrush        & \XSolidBrush            & 23.58                & -                     & \XSolidBrush        & \XSolidBrush       \\      
                         &Make-A-Story\cite{rahman2023make}     & 36.55               & 53.10                 & 23.74                & 52.08                 & \XSolidBrush        & \XSolidBrush                 \\  
                         &\CC{}AR-LDM\cite{pan2022synthesizing} & \CC{}23.59          & \CC{}39.70            &\CC{} 19.28           &\CC{} 43.32            &\CC{} \XSolidBrush   &\CC{} \XSolidBrush          \\ 
                         &Causal-Story\cite{song2023causal}     & -                   & -                     & 19.03                & -                     & \XSolidBrush        & \XSolidBrush               \\ 
                         &\CC{}StoryImager (Ours)               &\CC{}\textbf{22.27}  &\CC{}\textbf{36.51}    &\CC{}\textbf{18.32}   &\CC{}\textbf{35.33}    &\CC{}\textbf{18.11}  &\CC{}\textbf{34.20}         \\ 
\bottomrule
\end{tabular}}
\label{table1}
\vspace{-0.2cm}
\end{table}

\begin{table}[t] \small
\centering
\caption{The computational cost and time requirements of AR-LDM and our StoryImager on Pororo-SV dataset.}
\vspace{-0.4cm}
\resizebox{0.99\linewidth}{!}{
\begin{tabular}{lcccccccc}
\toprule
\multirow{2}*{Method}             & \multicolumn{4}{c}{Training}                                          & \multicolumn{4}{c}{Inference}                                     \\ 
                                  \cmidrule(lr){2-5}                                                       \cmidrule(lr){6-9}                                  
                                  &GPU-Memory      &Time           &Tasks                   &Models       & GPU-Memory  &Speed          &Tasks                   &Models      \\ \midrule              
AR-LDM \cite{pan2022synthesizing} &38G+38G         &52h+52h        &2                       &2            & 16G+16G     & 14.5s         &2                       &2           \\ 
StoryImager                       &\textbf{12G}    &\textbf{30h}   &\textbf{\textgreater2}  &\textbf{1}   &\textbf{8G}  &\textbf{8.0s}  &\textbf{\textgreater2}  &\textbf{1}   \\ 
\bottomrule
\end{tabular}}
\label{table2}
\vspace{-0.4cm}
\end{table}

\subsection{Qualitative Evaluation}

\begin{figure*}[t]
  \centering
  \includegraphics[width=\linewidth]{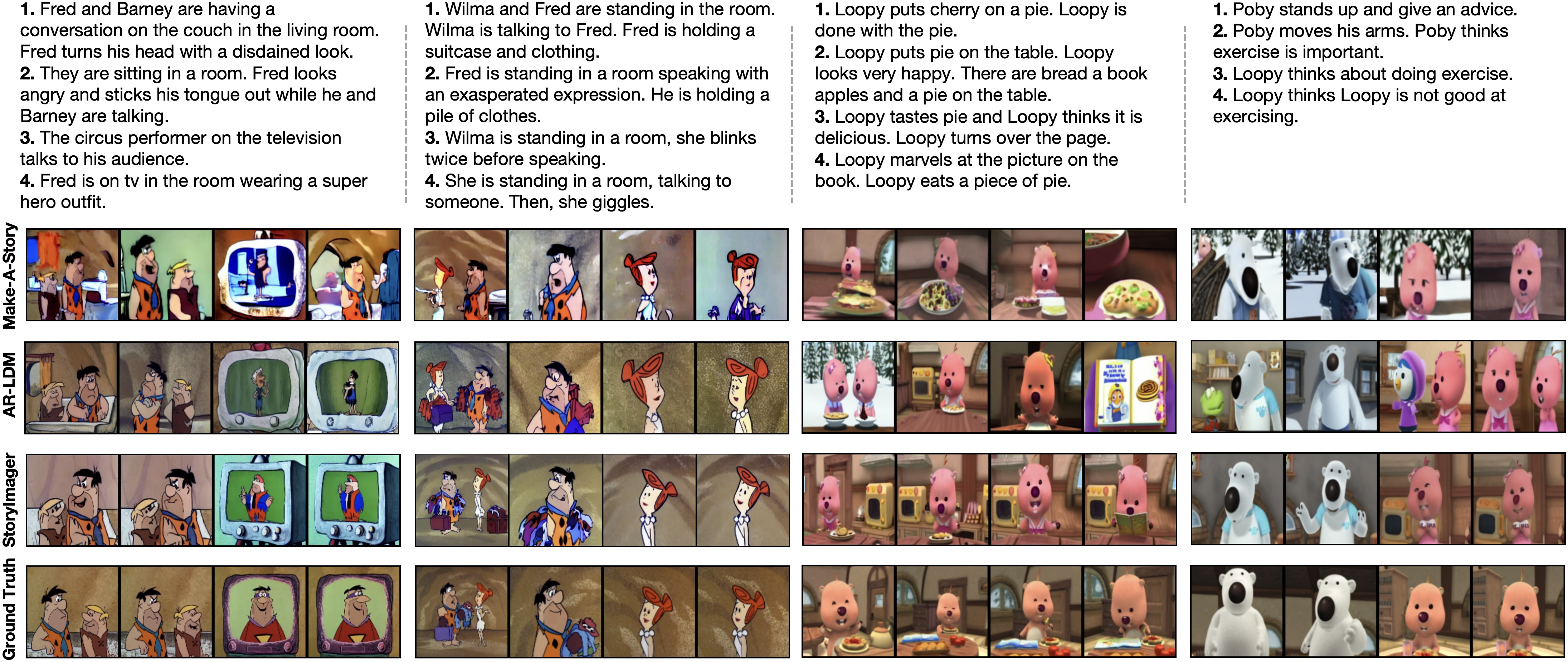}
  \caption{Comparison of story visualization results between Make-A-Story, AR-LDM, and our proposed StoryImager on Flintstones-SV and Pororo-SV datasets.}
  \label{fig5}
\vspace{-0.2cm}
\end{figure*}

\begin{figure*}[t]
  \centering
  \includegraphics[width=\linewidth]{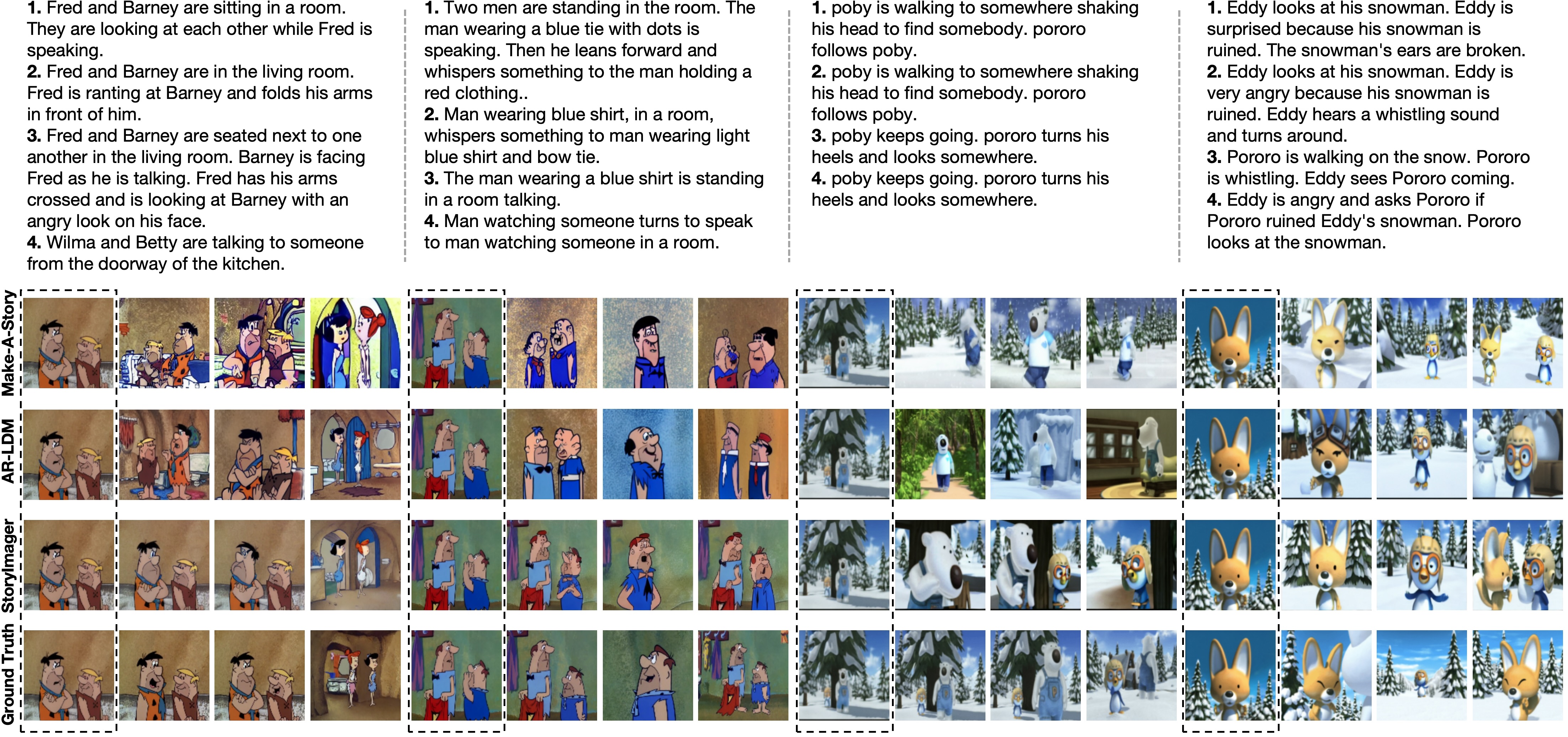}
  \caption{Comparison of story continuation results between Make-A-Story, AR-LDM, and our proposed StoryImager on Flintstones-SV and Pororo-SV datasets.}
  \label{fig6}
\vspace{-0.4cm}
\end{figure*}

Figures~\ref{fig5} and~\ref{fig6} show examples of visual comparisons between our StoryImager, AR-LDM\cite{pan2022synthesizing}, and Make-A-Story\cite{rahman2023make} on story visualization and continuation tasks, respectively. 
For story visualization, the characters and backgrounds synthesized by AR-LDM and Make-A-Story are not consistent with other frames.
As the synthesized story images are shown in Figure~\ref{fig5} of Pororo-SV, 
the appearance of ``Poby'' and the background of ``Loopy'' are inconsistent with other frames.
The same issue exists in the Flintstones-SV dataset as well, 
such as the disappearance of ``couch'' in the second frames and the transformation of the ``television''.
For story continuation, we also observed the inconsistent issue of AR-LDM and Make-A-Story. 
Additionally, there is a potential for the visual features from the first frame to be disregarded during the generation of subsequent frames.
Compared with AR-LDM and Make-A-Story, our StoryImager can synthesize contextual coherent images during story visualization
and incorporate the visual features from the first frame during the story continuation.

\begin{figure}[t]
  \centering
  \includegraphics[width=\linewidth]{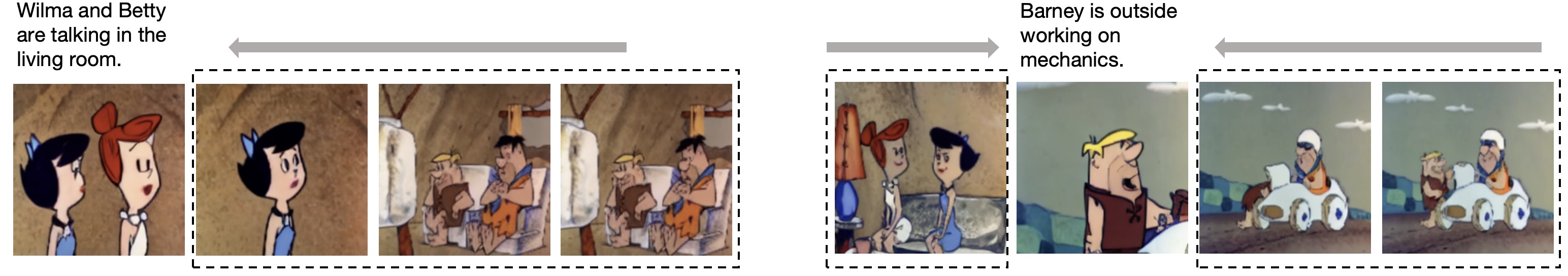}
    \vspace{-0.4cm}
  \caption{Examples of story backtracking and infilling synthesized by our StoryImager.}
  \label{fig7}
  \vspace{-0.4cm}
\end{figure}

Moreover, we show the story completion capability of StoryImager, as illustrated in Figure~\ref{fig7}. 
Our StoryImager is the first story completion model.
It can generate the first frame based on given frames and the textual description of the target frame, enabling story backtracking functionality. 
Additionally, our model supports adding new frames to the current story image, achieving story infilling.

\subsection{Human Evaluation}

We conducted the human evaluation to assess the Visual Quality (VQ), Visual Consistency (VC), and Story Relevance (SR) of the generated results. 
For each story, we had 12 evaluators rank the synthesized image sequences from our model and the two other top-performing models \cite{rahman2023make, pan2022synthesizing} on a scale of 1 to 3.
We provide ground truth as a reference.
Each model generated 300 story image sequences for each dataset and task. 
As shown in Table~\ref{table3}, our model achieves the best performance in terms of all these three criteria.
Owing to the powerful storyboard-Gen and decomposed Frame-Story CAM, StoryImager significantly outperforms other models in Visual Consistency (VC).

\begin{table}[t] \small
\centering
\caption{Human evaluation results of story visualization and continuation tasks on Pororo-SV and Flintstones-SV datasets.}
\vspace{-0.4cm}
\resizebox{\linewidth}{!}{
\begin{tabular}{lccccccc}
\toprule
\multirow{2}*{Dataset}                       &\multirow{2}*{Criterion(Avg)}   & \multicolumn{3}{c}{Visualization}                            & \multicolumn{3}{c}{Continuation}          \\ 
                                                                         \cmidrule(lr){3-5}                                             \cmidrule(lr){6-8}                           
                                             &      &Make-A-Story\cite{rahman2023make} &AR-LDM\cite{pan2022synthesizing}  &Ours  &Make-A-Story\cite{rahman2023make}   &AR-LDM\cite{pan2022synthesizing}    &Ours   \\   \midrule              
\multirow{3}*{Pororo-SV}                     &VQ $\downarrow$            &2.40       &1.89          &\textbf{1.71}          &2.35              &1.87                &\textbf{1.78}       \\ 
                                             &VC $\downarrow$            &2.63       &1.92          &\textbf{1.45}          &2.60              &1.85                &\textbf{1.55}       \\ 
                                             &SR $\downarrow$            &2.52       &1.82          &\textbf{1.66}          &2.48              &1.80                &\textbf{1.72}      \\ 
                                             & \\[-2.3ex]       
                                             \hline        
\multirow{3}*{Flintstones-SV}                        & \\[-2.3ex]       
                                             &VQ $\downarrow$            &2.45      &1.85           &\textbf{1.70}          &2.39              &1.86               &\textbf{1.75}         \\  
                                             &VC $\downarrow$            &2.74      &1.89           &\textbf{1.37}          &2.70              &1.82               &\textbf{1.48}         \\ 
                                             &SR $\downarrow$            &2.66      &1.75           &\textbf{1.59}          &2.61              &1.74               &\textbf{1.65}       \\ 
\bottomrule
\end{tabular}}
\label{table3}
\vspace{-0.2cm}
\end{table}

\subsection{Ablation Study}

To verify the effectiveness of different components in the proposed StoryImager, we conduct ablation studies on the story visualization and continuation tasks, respectively. 
The results of Pororo-SV and Flintstones-SV are shown in Table~\ref{table4}.
The components being evaluated include Contextual Feature Extractor (Contextual-FE), and Frame-Story Cross Attention Module (Frame-Story CAM). 
Our baseline is a modified Stable diffusion \cite{rombach2022high} adapted for story visualization and continuation tasks with a target masking strategy. 
The baseline concatenates all the story captions into a single long text.

From Table~\ref{table4}, we observe that the proposed Contextual-FE can significantly reduce FID and FSD on these two tasks.
If we further introduce the Frame-Story CAM to decompose the cross attention, we can observe a further improvement of FID and FSD.
% Armed with Intra-SI, the model also decreases FID and FSD from 24.39 and 26.56 to 19.27 and 20.19 for the story visualization task on Pororo.
% The proposed Context-aware SVF can decrease FID and FSD from 19.27 and 20.19 to 15.75 and 17.45.
The ablation studies demonstrate the effectiveness of our proposed modules in both story visualization and completion tasks.

\begin{table}[t] 
  \centering
  \caption{The performance of different components of our model on the test set of Pororo-SV and Flintstones-SV.}
  \vspace{-0.4cm}
    \resizebox{0.9\linewidth}{!}{ %< auto-adjusts font size to fill line
  \begin{tabular}{llcccc}
  \toprule
  \multirow{2}*{Task}     &  \multirow{2}*{Method}       & \multicolumn{2}{c}{Pororo-SV}                               & \multicolumn{2}{c}{Flintstones-SV}                         \\ 
  \cmidrule(lr){3-4} \cmidrule(lr){5-6}    
                          &                              & FID ($\downarrow$)     & FSD ($\downarrow$)       & FID ($\downarrow$)       & FSD ($\downarrow$)                         \\ \midrule
                          &Baseline                      & 21.14                  & 41.18                    & 28.19                    & 42.78                                    \\
  Story Visualization           &+ Contextual-FE               & 18.02                  & 36.34                    & 25.75                    & 39.56                                         \\            
                          &+ Frame-Story CAM             & \textbf{15.63}         & \textbf{28.13}           & \textbf{22.27}           & \textbf{36.51}                          \\ \hline      
                          &Baseline                      & 20.65                  & 40.10                    & 23.13                    & 41.15                                         \\
  Story Completion              &+ Contextual-FE               & 18.06                  & 34.19                    & 21.82                    & 37.88                                       \\   
                          &+ Frame-Story CAM             & \textbf{14.72}         & \textbf{26.77}           & \textbf{18.11}           & \textbf{34.20} \\ 
  \bottomrule
  \end{tabular}}
  \label{table4}
  %\vspace{-0.4cm}
  \end{table}

\section{Conclusion}

In this paper, we propose a novel unified and contextual coherent framework for the story visualization and completion model, namely StoryImager.
The StoryImager inherits the storyboard generative ability of a large pre-trained text-to-image model and extends the story image generation task. 
Specifically, we introduce a Target Frame Masking Strategy to unify different story image generation tasks.
Furthermore, we propose a Frame-Story Cross Attention Module that decomposes the cross attention for local frame fidelity and global storyboard conherence.
Moreover, we design a Contextual Feature Extractor, which extracts global context information and synthesizes context visual features.
Our StoryImager achieves significant improvements on two challenging datasets.
Compared to previous models, our StoryImager can synthesize more coherent story images while reducing hardware and time requirements.

\clearpage  % TODO REVIEW/FINAL: This \clearpage needs to be removed from both review and camera-ready versions.

\bibliographystyle{splncs04}
\bibliography{main}
\end{document}